\documentclass{article}

 \usepackage[dblblindworkshop, final]{neurips_2025}
\workshoptitle{Mech. Interp.}

\usepackage{float}
\usepackage{algorithm}
\usepackage{algpseudocode}
\usepackage{amsmath}
\usepackage{graphicx}
\usepackage{subcaption}
\usepackage[utf8]{inputenc} 
\usepackage[T1]{fontenc}    
\usepackage{hyperref}       
\usepackage{url}            
\usepackage{booktabs}       
\usepackage{amsfonts}       
\usepackage{nicefrac}       
\usepackage{microtype}      
\usepackage{xcolor}         

\title{Automatically Finding Rule-Based Neurons\\in OthelloGPT}

\hypersetup{
    colorlinks=true,
    linkcolor=blue,
    urlcolor=blue,
}

\author{%
  Aditya Singh\thanks{Equal contribution} \\
  University of Chicago\\
  singh.adityak1@gmail.com \\
  \And
  Zihang Wen\footnotemark[1] \\
  Carnegie Mellon University \\
  \texttt{zihangw@cs.cmu.edu} \\
  \AND
  Srujananjali Medicherla\footnotemark[1] \\
  Independent \\
  \texttt{srujananjali888@gmail.com} \\
  \And
  Adam Karvonen \\
  Independent \\
  \texttt{adam.karvonen@gmail.com} \\
  \And
  Can Rager \\
  Independent \\
  \texttt{canrager@gmail.com} \\
}

\begin{document}

\maketitle

\begin{abstract}
OthelloGPT, a transformer trained to predict valid moves in Othello, provides an ideal testbed for interpretability research. The model is complex enough to exhibit rich computational patterns, yet grounded in rule-based game logic that enables meaningful reverse-engineering.  We present an automated approach based on decision trees to identify and interpret MLP neurons that encode rule-based game logic. Our method trains regression decision trees to map board states to neuron activations, then extracts decision paths where neurons are highly active to convert them into human-readable logical forms. These descriptions reveal highly interpretable patterns; for instance, neurons that specifically detect when diagonal moves become legal. Our findings suggest that roughly half of the neurons in layer 5 can be accurately described by compact, rule-based decision trees ($R^2 > 0.7$ for 913 of 2,048 neurons), while the remainder likely participate in more distributed or non-rule-based computations. We verify the causal relevance of patterns identified by our decision trees through targeted interventions. For a specific square, for specific game patterns, we ablate neurons corresponding to those patterns and find an approximately 5-10 fold stronger degradation in the model's ability to predict legal moves along those patterns compared to control patterns. To facilitate future work, we provide a Python tool that maps rule-based game behaviors to their implementing neurons, serving as a resource for researchers to test whether their interpretability methods recover meaningful computational structures.
\end{abstract}


\section{Introduction}
A long-standing goal of interpretability research is to construct human-interpretable replacement models that explain model behavior in terms of explicit representations and computations. Prior work has demonstrated the feasibility of constructing models with fully known internal mechanisms \citep{lindner_tracr_2023}, or has attempted to construct replacement models for modern LLMs \citep{ameisen2025circuit}. However, handcrafting models does not reveal how naturally trained networks develop to solve tasks, while evaluating replacement models for modern LLMs can be difficult due to a lack of ground truth features \citep{karvonen_measuring_2024}.

We propose using OthelloGPT, a transformer trained to predict valid moves in Othello, as an interpretability testbed for uncovering ground truth behavior in neural networks. While substantially more complex than existing toy models \citep{elhage2021mathematical, elhage2022superposition, lindner_tracr_2023}, OthelloGPT remains grounded in rule-based game logic that enables meaningful reverse-engineering. In particular, prior work \citep{lin_othellogpt_nodate} suggests that certain neurons in OthelloGPT implement rule-based behavior, such as neurons responding to a specific diagonal pattern, which provides a concrete target for reverse engineering.

\begin{figure}[!htbp]
  \begin{center}
    \includegraphics[width=1\textwidth]{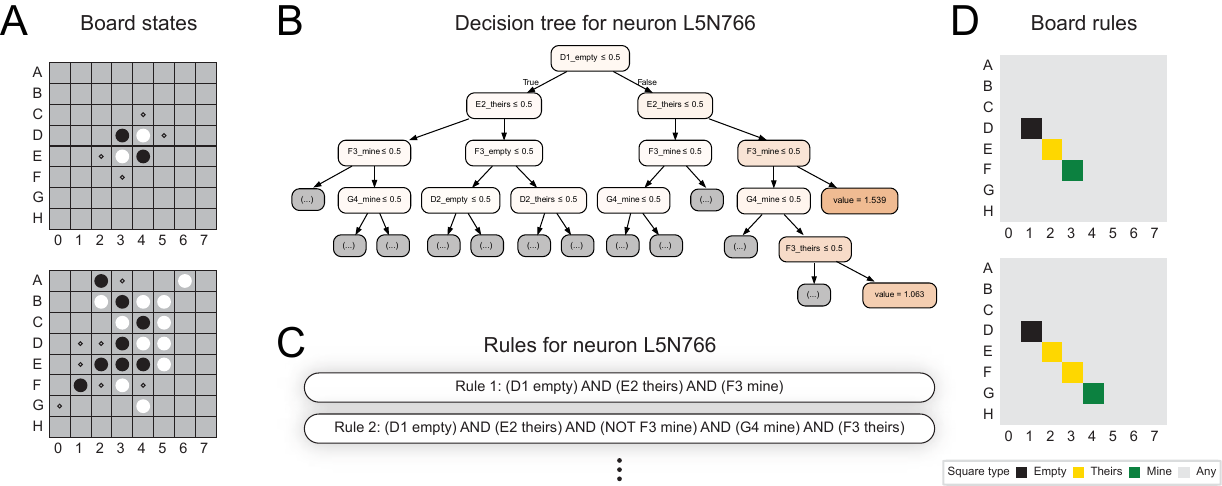}
  \end{center}
  \caption{Overview of neuron interpretation pipeline. Board state features are used as training data for decision trees to predict neuron activations (\textbf{A}, \textbf{B}). Note: small circles in \textbf{A} and \textbf{B} are all possible legal moves. High activation decision paths are then extracted to obtain logical rules for each neuron (\textbf{C}, \textbf{D}). We provide an interactive visualization of decision trees in this \href{https://colab.research.google.com/drive/1kKLj9c3elB0yjJoBZnkHqFl3ZWygP6zw}{Colab Notebook}.}\label{Fig:dt-training}
\end{figure}

We automate the process of finding rule-based neurons in OthelloGPT by training regression decision trees that map board state features to neuron activations, and then extract decision paths where neurons are highly active to convert neurons into human-readable logical forms (Figure \ref{Fig:dt-training}). The result is a disjunctive normal form (DNF; OR-of-ANDs) description for each neuron, where each AND corresponds to a high activation decision path of the neuron's decision tree. We may then specify a rule-base game query (an AND of features), and automatically surface the implementing neurons by evaluating if their DNF evaluates to True when given the query.

Through traditional machine learning metrics such as $R^2$ scores, metrics specific to OthelloGPT, as well as causal interventions, we verify that our decision trees are effective at predicting neuron activations, that they pick up on the relevant features, and that their decision paths are mechanistically faithful. To facilitate future work, we provide a Python tool that maps rule-based game behaviors to their implementing neurons, allowing others to test if their interpretability methods can recover the same ground truth structure as our decision trees.

Our contributions are summarized as follows:
\begin{enumerate}
    \item An automatic method for discovering rule-based neurons in OthelloGPT. We validate our method through traditional machine learning metrics and causal interventions.
    \item A Python tool mapping rule-base game behaviors to their implementing neurons.
\end{enumerate}

We release code and decision trees at \url{https://github.com/zihangwen/OthelloReverseEngineering}. The colab is available at \url{https://colab.research.google.com/drive/1kKLj9c3elB0yjJoBZnkHqFl3ZWygP6zw?usp=sharing}.

\section{Background}
\subsection{Othello-GPT}
OthelloGPT is a transformer with 25M parameters trained to predict legal moves in the board game Othello \citep{li_emergent}. By training with the standard autoregressive loss over sequences of random legal moves, OthelloGPT learns to predict a uniform distribution over legal moves. Although the model receives no explicit knowledge of game rules, it still achieves over 99\% legal move accuracy. Prior work interpreting OthelloGPT uncovered that the model learns to represent a linear world model of the Othello board in its internal states \citep{li_emergent, nanda_emergent_2023}. The model then follows a two-step procedure for predicting legal moves, whereby it first updates its world model of the board state, and then uses this world model to predict legal moves \citep{li_emergent, nanda_emergent_2023, hazineh2023linear, karvonen_measuring_2024, robert_aizi_research_nodate, lin_othellogpt_nodate, jmaar_exploring_nodate}.

\subsection{Linear world model}
\cite{li_emergent} demonstrated that Othello-GPT develops internal world models that track board states during gameplay. Their initial investigations suggested these representations were non-linear. When probing for absolute tile colors \{BLACK, WHITE, EMPTY\}, linear probes achieved poor performance with error rates of more than 20\% across layers, while non-linear probes (2-layer MLPs) dramatically outperformed them, achieving error rates as low as 1.7-4.6\% in deeper layers.

\cite{nanda_emergent_2023} extended this finding by shifting to player-relative classification \{MINE, YOURS, EMPTY\} and showed that the internal representation could be extracted using linear probes, which achieved a remarkable accuracy exceeding 99\% from the end of layer four onwards. Intervening on these probe directions led to causal changes in model behavior. In addition to linearly representing the board state, \cite{nanda_emergent_2023} found that Othello-GPT linearly represents which tiles are being flipped at each timestep by training a probe to classify between FLIPPED vs. NOT-FLIPPED states. Follow-up work by \cite{hazineh2023linear} further validated these findings across different model scales, showing that linear world representations emerge even in minimal transformers with just one layer and attention head, and that these representations grow stronger with model depth. 

As a way to evaluate if our decision trees are capturing the relevant features, we construct a metric based on board state probes, flipped probes, as well as our own probes for if a square was just played (Section \ref{metric-results}).

\subsection{Individual Decision Rules}
Analysis in \cite{lin_othellogpt_nodate} revealed that many neurons in Othello-GPT follow explicit logical rules with clear conditional structure. For example, they find the MLP neuron L1N421 represents the decision rule ``If the move A4 was just played AND B4 is occupied AND C4 is occupied $\Rightarrow$ update B4+C4+D4 to THEIRS''.

More broadly, \cite{lin_othellogpt_nodate} identified specialized MLP neurons in layers 4-6 that act as classifiers for specific board patterns that make particular moves legal, and verified through causal interventions that these neurons were reponsible for implementing legal move prediction along these patterns. Our work can be seen as a direct extension, where we attempt to full reverse-engineer rule-based behavior in OthelloGPT.

\section{Identifying Rule-based Neurons}
\subsection{Decision Trees}
Decision trees present a natural machine learning technique for discovering rule-based neurons, as they are inherently rule-based themselves. Hence, if a neuron can be well-explained by a decision tree, this can tell us whether or not a neuron's firing pattern follows logical rules. 

We explore two variants of decision trees: regressor decision trees, which directly predict a neuron's activation value, and binary decision trees, which predict whether a neuron is \emph{on} or \emph{off}. For binary decision trees, we define \emph{on} as the neuron's activation is greater than 0.1 of its max activation (over the train set), and \emph{off} otherwise. In both cases, the input features used to train the trees are as the MINE/YOURS/EMPTY status of each square on the board ($64\times3$), the most recent move ($64$), and whether a tile has been flipped in the most recent move ($64$).

We train depth $4$ decision trees over 6000 games, regularized with a minimum node split count of $100$ and a minimum leaf node count of $50$. In addition to using ground truth board state information, we also explored training regression trees using model internal board state information as inputs. Specifically projections along probe directions as features, and found them to be similarly performant to the trees trained on ground truth features. For simplicity, we use the ground truth feature decision trees for all our following experiments.

\subsection{Baselines}
We explore the following simpler alternatives to decision trees for identifying rule-based neurons. In all cases, the input features are the same game state features as in the training of the decision trees:
\begin{enumerate}
    \item Lasso (L1) regression. We train linear regression with an L1 penalty on a single layer, encouraging sparsity so that only a small subset of input features is used to predict each neuron’s activation. We interpret the top-weighted coefficients as the key features influencing the neuron’s behavior. We compare lasso regression $R^2$ scores to decision tree regression $R^2$ scores.
    
    \item RIPPER. We use RIPPER (\textit{Repeated Incremental Pruning to Produce Error Reduction}), a classic rule-based learner for binary classification that induces compact conjunctive rules to predict whether a neuron is \emph{on} or \emph{off} (same “on’’ definition as in the binary decision-tree experiments). We evaluate each neuron’s classifier by its F1-score on the \emph{on} class and compare it with the binary decision-tree F1-scores. For containment analysis, top-$k$ features are extracted from the learned rules using a scoring-and-filtering procedure detailed in Appendix~\ref{app:ripper_features}.
\end{enumerate}

\subsection{Results}\label{metric-results}
\begin{figure}[!htbp]
  \begin{center}
    \includegraphics[width=1\textwidth]{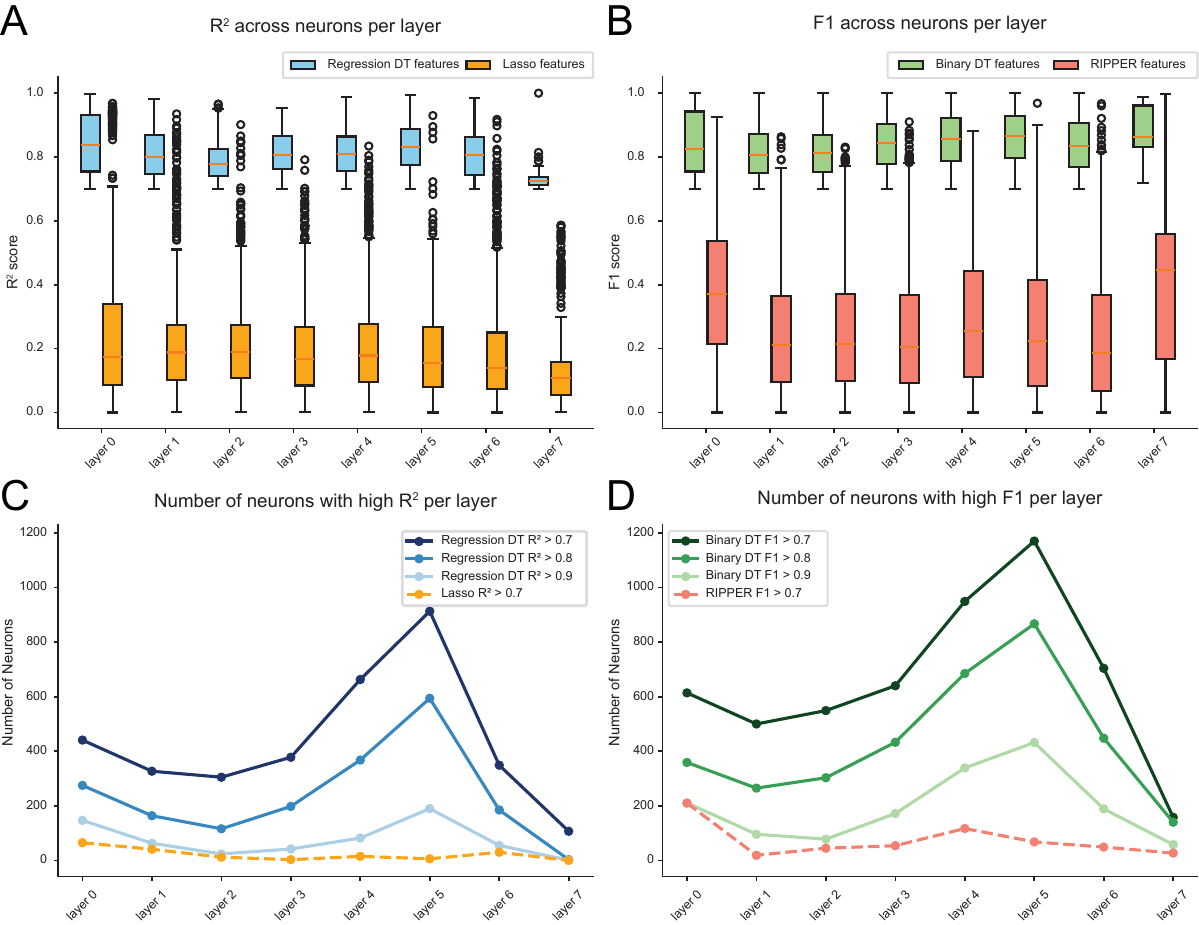}
  \end{center}
  \caption{Comparison of decision trees to other baselines. \textbf{A} We evaluate regression methods on $R^2$ scores and \textbf{B} classification methods on $F1$ scores. \textbf{C}, \textbf{D}: We show the number of interpretable rule-based neurons with score cutoffs of 0.7, 0.8, and 0.9}\label{Fig:test-scores}
\end{figure}

We find that the simpler baselines consistently underperform compared to the decision trees on both $R^2$ and F1 scores (Figure \ref{Fig:test-scores} \textbf{A}, \textbf{B}). We also show the number of interpretable rule-based neurons with cutoffs of 0.7, 0.8, and 0.9 of the scores (Figure \ref{Fig:test-scores} \textbf{C}, \textbf{D}). These show that ``valid-move'' neurons in layers 5 and 6 are more interpretable with rules.

\begin{figure}[!htbp]
  \begin{center}
    \includegraphics[width=1\textwidth]{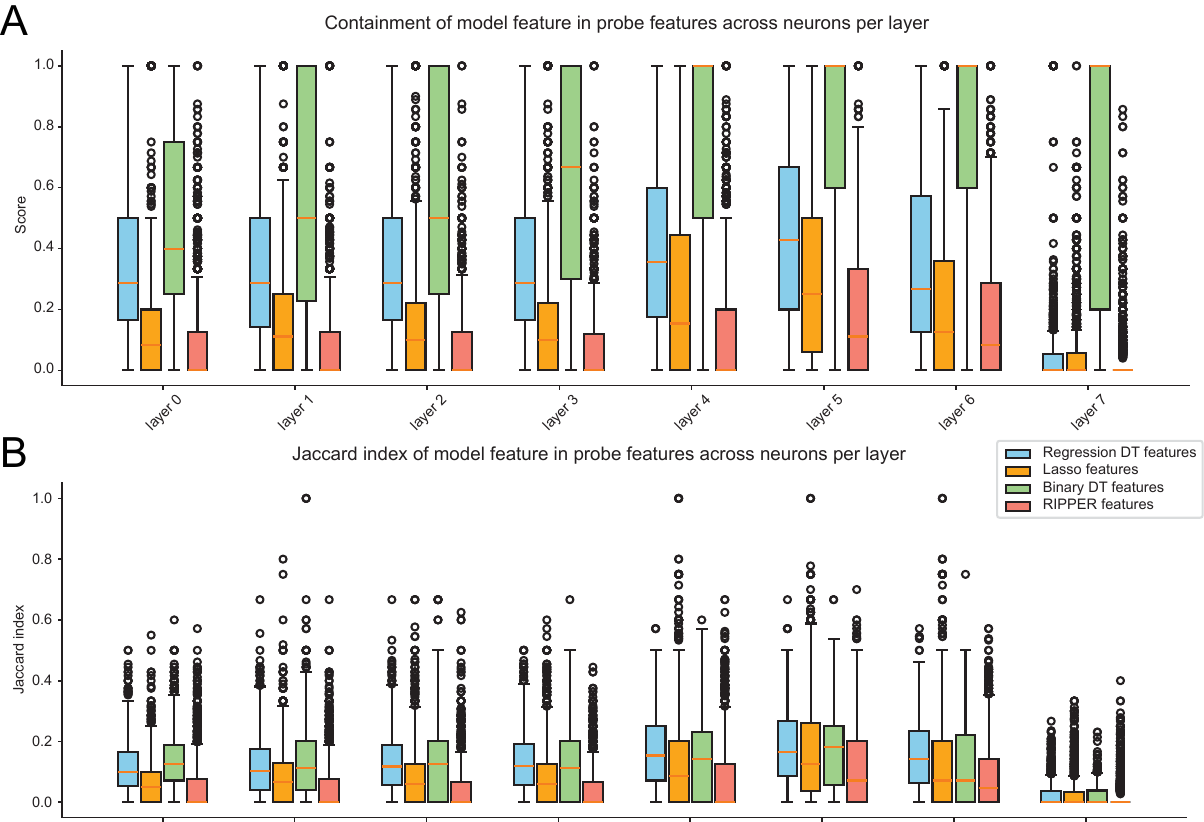}
  \end{center}
  \caption{We evaluate all methods on two contrastive probe feature metrics. A: containment metric. B: Jaccard metric.}\label{Fig:contrastive-metrics}
\end{figure}

To further assess how well the models recover meaningful input features, we evaluated them using two contrastive probe metrics: (i) containment, measuring the fraction of each model’s top-ranked features that overlap with features identified via weight-based direct linear attribution (DLA) using board-state probes (Appendix ~\ref{probe_feature_extraction}), and (ii) the Jaccard index, quantifying overall similarity between the top-ranked feature sets and the probe feature set (Figure \ref{Fig:contrastive-metrics}).

\paragraph{Containment metric.}
Given a method’s selected top-$k$ features $F_{\mathrm{method}}$ for each neuron, and the set of probe-identified features $F_{\mathrm{probe}}$, we define
\begin{equation}
    C(F_{\mathrm{method}},F_{\mathrm{probe}}) = \frac{|F_{\mathrm{method}} \cap F_{\mathrm{probe}}|}
         {|F_{\mathrm{probe}}|}
\end{equation}
\paragraph{Jaccard index.}
The Jaccard index measures similarity between finite non-empty sample sets. It is defined as the size of the intersection divided by the size of the union of the sample sets:
\begin{equation}
    J(F_{\mathrm{method}},F_{\mathrm{probe}})=\frac{\left|F_{\mathrm{method}}\cap F_{\mathrm{probe}}\right|}{\left|F_{\mathrm{method}}\cup F_{\mathrm{probe}}\right|}
\end{equation}

We treat these probe-derived features as a reference ground truth, following prior work and our own analyses confirming their stability and distinctiveness. The intuition behind these metrics is that we can “read off” which squares a neuron attends to by examining the linear probe weights; we then check whether our trained models highlight the same input signals. Across all metrics, including $R^2$, $F1$, containment, and Jaccard, we found decision trees to be the most performant.

\subsection{Automatically surfacing implementing neurons}\label{neuron_search}
Given a rule-based query, such as ``C0 is blank AND D1 is theirs AND E2 is mine'', we aim to use our decision trees to automatically surface neurons that fire for this query. 

To do so, we use each neuron's decision tree to associate it with a disjunctive normal form (DNF), and then evaluate whether the neuron's DNF evaluates to True when given the rule-based query. In more detail, for each neuron we first separate out the "on" leaf nodes and "off" leaf nodes. This is already given for the binary decision trees, while for the regression decision trees we use Otsu's threshold, which sorts data into two groups such that the within group variance is minimized. The decision paths for the "on" leaf nodes then correspond to the AND clauses of the neuron's DNF.  

To make the rules easier to read, we rewrite and simplify them. For example, a decision tree might separately check “E2 is theirs” and, if not, “E2 is empty,” without ever explicitly concluding “E2 is mine.” By combining the conditions “NOT E2 is theirs” and “NOT E2 is empty,” we can express the same logic more clearly as “E2 is mine.” Similarly, when a rule redundantly includes both “E2 is theirs” and “NOT E2 is empty,” we keep only the stronger positive condition and simplify it to “E2 is theirs.” This process of merging fragmented conditions into explicit statements transforms messy, machine-generated decision paths into clear, human-readable rules. Figure \ref{Fig:dt-training} \textbf{B}, \textbf{C} shows an example of a decision tree and its corresponding generated rules.


\section{Intervening on Rule-based Game Behaviors}
\subsection{Layer-wise rule-based neuron intervention}
\begin{figure}[!htbp]
  \begin{center}
    \includegraphics[width=1\textwidth]{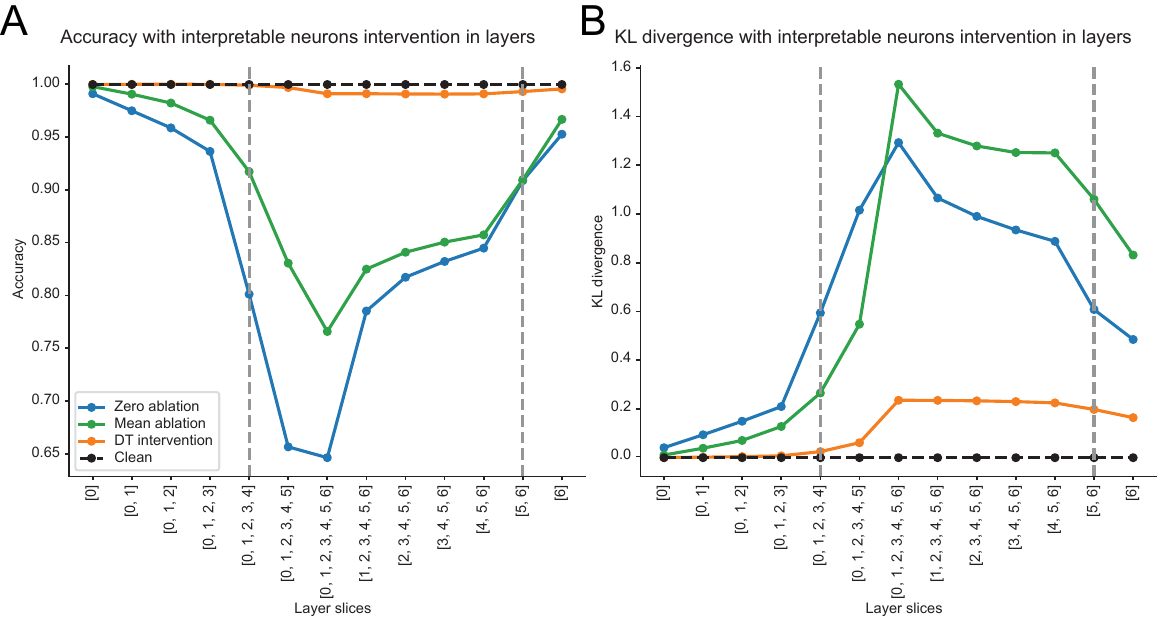}
  \end{center}
  \caption{Layer-wise rule-based neuron (cutoff of 0.7) intervention. \textbf{A}: Accuracy of intervention. \textbf{B}: KL divergence between before and after intervention. The dashed lines show the ablation of board-state related layers ([0,1,2,3,4]) and valid-move related layers ([5,6]) from the previous literature.}\label{Fig:intervention-layer}
\end{figure}

We used a cutoff of 0.7 to select interpretable, rule-based neurons. Our first intervention study involved replacing MLP neurons with regression decision tree neurons in entire layers (Figure \ref{Fig:intervention-layer}). Specifically, we replaced them in the order of the first N layers and the last N layers. (We excluded layer 7 because it appears to perform final output cleaning and is not easily explained by simple rules.) After substituting the selected neurons with the outputs of their corresponding decision tree models, the network’s valid-move prediction accuracy remained high, and the KL divergence of the output logits stayed low. These results indicate that the decision trees capture the essential functional behavior of the original neurons.

In contrast, completely removing these interpretable neurons or replacing them with mean activations caused a sharp drop in OthelloGPT’s performance. This demonstrates that the identified neurons carry critical information and are causally important for OthelloGPT's computation. The degradation is especially pronounced when intervening on layers 5 and 6, consistent with prior findings that these layers play a key role in rule-based valid move prediction.

\subsection{Fine-grained interventions}
\begin{figure}[!htbp]
  \begin{center}
    \includegraphics[width=0.85\textwidth]{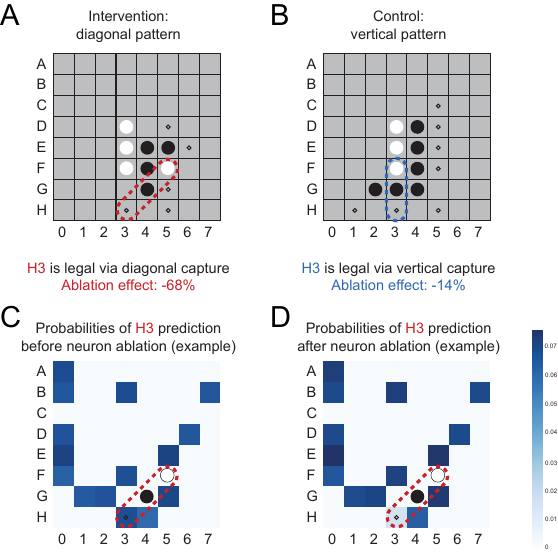}
  \end{center}
  \caption{Fine-grained intervention. \textbf{A}: Intervention pattern example: valid move via diagonal pattern. \textbf{B}: Control pattern example: valid move via diagonal pattern. Note: small circles in \textbf{A} and \textbf{B} are all possible legal moves. \textbf{C}, \textbf{D}: Output Probabilities of OthelloGPT output on a game example where H3 is legal via diagonal pattern. \textbf{C}: Before intervention. \textbf{D}: After intervention, the probability for the intervened square H3 drops to in between the legal and illegal square probabilities. Run more interventions in the \href{https://colab.research.google.com/drive/1kKLj9c3elB0yjJoBZnkHqFl3ZWygP6zw}{Colab Notebook.}}\label{Fig:intervention-setup}
\end{figure}

Suppose a decision tree claims that a neuron fires when a particular square is legal along a certain diagonal. If this is indeed the case, then we should expect that for positions where that move is legal due to the diagonal, ablating the neuron reduces the model's probability for that square. Moreover, this should not greatly reduce the model's probability for that square when it is legal due to a different pattern, such as a certain row.

We perform an intervention experiment along these lines for each of the 60 squares that can be played in OthelloGPT. For each square, we specify an intervention condition and a control condition, both of which are distinct 3 square patterns that make the given square legal. A visual depiction is shown in (Figure \ref{Fig:intervention-setup} \textbf{A}, \textbf{B}). We then use our pipeline from Section \ref{neuron_search} to collect the neurons that respond to the intervention pattern. Over a 500 game test set, we filter for positions where the intervention condition is satisfied but not the control condition, as well as for positions where the control condition is satisfied but not the intervention condition. We then zero ablate the intervention condition neurons over both sets. Further details about the experimental setup and the steps taken to avoid confounding are provided in Appendix \ref{sec:intervention_confounding}.

\subsubsection{Metrics}
We include natural metrics such as probability difference and logit difference. While another natural metric to include would be the change in legal move accuracy (defined as inaccurate if and only if an illegal square is in the top $K$ logits, where $K$ is the number of legal moves in the given position), OthelloGPT's prediction of a uniform distribution over legal moves complicates this metric. In particular, since each legal move receives probability $1/K$ (where $K$ is the number of legal moves), and each illegal move receives probability $0$, it is difficult to completely move the intervention legal square out of the top $K$. The situation is visually depicted in Figure \ref{Fig:intervention-setup} \textbf{C}, \textbf{D} (observe the shading for square H3).
Hence, we instead include the following three metrics: dropping the intervention square probability below 1\%, 5\%, and 10\% of its original probability.

\subsubsection{Results}

Averaging results over all 60 squares, we find that the legal square probability drops nearly twice as much in the intervention conditions as in the control conditions (Table~\ref{tab:intervention_control}). For the below 1\% of its original probability metric, we find $\sim$4x greater effect in the intervention than control conditions.

We found that intervention results for the $12$ squares adjacent to the middle 2x2 region were subpar, suggesting that there is greater shared behavior among such squares. Hence, we perform a variant where we average only over the 48 squares outside the middle 4x4. As shown in Table~\ref{tab:intervention_control}, we find substantial improvement in our metrics, most notably with a $\sim$10x greater effect in the intervention than control conditions for the below 1\% of original probability metric (0.077 vs. 0.0076).
We interpret these results as evidence that the neurons causally implement the behaviors their decision trees describe, but note that the incomplete effect on model behavior suggests that there are other prediction mechanisms that we are missing.

\begin{table}[htbp]
\centering
\caption{Intervention vs. Control Condition Results}
\label{tab:intervention_control}
\begin{tabular}{lcccc}
\toprule
\textbf{Metric} & \multicolumn{2}{c}{\textbf{All 60 Squares}} & \multicolumn{2}{c}{\textbf{48 Squares (excl. middle 4x4)}} \\
\cmidrule(lr){2-3} \cmidrule(lr){4-5}
 & \textbf{Intervention} & \textbf{Control} & \textbf{Intervention} & \textbf{Control} \\
\midrule
Logit Difference & 1.612 & 0.723 & 1.854 & 0.682 \\
Probability Difference & 0.057 & 0.031 & 0.062 & 0.031 \\
Clean Accuracy & 0.99996 & 0.99999 & 0.99995 & 0.99998 \\
Corrupted Accuracy & 0.994 & 0.997 & 0.998 & 0.999 \\
Accuracy Difference & 0.0059 & 0.0030 & 0.0016 & 0.00065 \\
Below 1\% & 0.058 & 0.015 & 0.077 & 0.0076 \\
Below 5\% & 0.191 & 0.062 & 0.239 & 0.040 \\
Below 10\% & 0.287 & 0.094 & 0.344 & 0.079 \\
\bottomrule
\end{tabular}
\end{table}

\section{Discussion} 




Our results demonstrate that a substantial fraction of neurons in OthelloGPT can be modeled as rule-based units whose activations follow explicit, human-interpretable logical conditions. Decision trees proved to be the most performant method across all quantitative metrics, outperforming both sparse linear models and classical rule learners. This suggests that many neurons in OthelloGPT implement discrete decision boundaries that are well approximated by a small set of conjunctive conditions.

Layer-wise dynamics and partial coverage. Consistent with prior work, interpretable rule-based neurons are concentrated in layers 5–6, while earlier layers (0–4) appear to build distributed board-state representations. Intervening on layers 5–6 yields the largest causal degradation in valid-move prediction, and fine-grained ablations show direction-specific drops in probability for the targeted legal moves. Yet replacing these neurons with their decision-tree surrogates preserves most accuracy, whereas ablating them harms specific sub-behaviors, indicating that OthelloGPT combines rule-like components with more continuous, distributed mechanisms that our trees only partially capture.

Conceptually, decision trees provide an advantage over linear probes or direct weight attribution: rather than assigning scalar importance to individual input features, they capture compositional logic ``ORs of ANDs'' that mirrors the conditional dependencies present in the game’s rules. This structured representation yields clearer hypotheses for mechanistic testing and helps to explain network behavior.

\section{Limitations}
We focus only on reverse-engineering rule-based behavior in OthelloGPT, which provides an incomplete picture of its internal computation. For example, \citet{jmaar_exploring_nodate} identify global “flip” circuits that propagate ownership changes across timesteps—mechanisms that are explicitly \textit{not} rule-based and thus fall outside the scope of our analysis. Our decision trees are also imperfect approximations: a neuron whose activation depends on long conjunctive dependencies (e.g., a length-5 AND clause) or on continuous features may not be well captured by the depth-4 trees we train.

Furthermore, our approach assumes that individual neurons are the natural unit of analysis, but OthelloGPT may implement higher-order computations distributed across neuron groups or attention heads. In these cases, decision trees trained per neuron may miss relational patterns or mixed features that only emerge jointly. Extending tree-based analysis to multi-neuron subspaces, deeper logical dependencies, or time-coupled features could help close these gaps.

\section{Conclusion}

We introduced a decision-tree–based framework for automatically identifying and interpreting rule-based neurons in OthelloGPT. Using both predictive metrics and causal interventions, we showed that many neurons follow compact logical rules that align with game mechanics, and that intervening on these neurons predictably alters model behavior. These results reinforce the view that OthelloGPT performs structured, rule-like reasoning grounded in an internal linear world model, while also revealing that such reasoning coexists with more distributed, continuous computation.

Our open-source tool and neuron–rule mapping provide a reproducible benchmark for testing interpretability methods against a model with known ground-truth structure. We hope this work advances the broader goal of connecting feature-level analyses, such as decision trees, probes, and sparse autoencoders, into a framework for mechanistically understanding how transformers implement reasoning computation.

\section*{Acknowledgements}
This collaboration is part of the Algoverse program. We especially thank Sunishchal Dev, Kevin Zhu for paper feedback and running the program and Callum McDougall for helpful mentorship.

\clearpage
\newpage

\bibliography{reference.bib}

\newpage
\appendix

\section{Decision tree scaling law}\label{dt_dataset}
We trained binary ground-truth classification decision trees on datasets of varying sizes (60, 600, 6,000, 12,000, and 30,000 games) and evaluated them on a 500-game test set. We found that performance almost stabilized once the training set reached about 6,000 games, with only marginal gains beyond that point. For computational efficiency, we therefore used the 6,000-game dataset for more complex decision tree variants, such as regression trees and probe-feature trees.

\begin{figure}[!htbp]
  \begin{center}
    \includegraphics[width=1\textwidth]{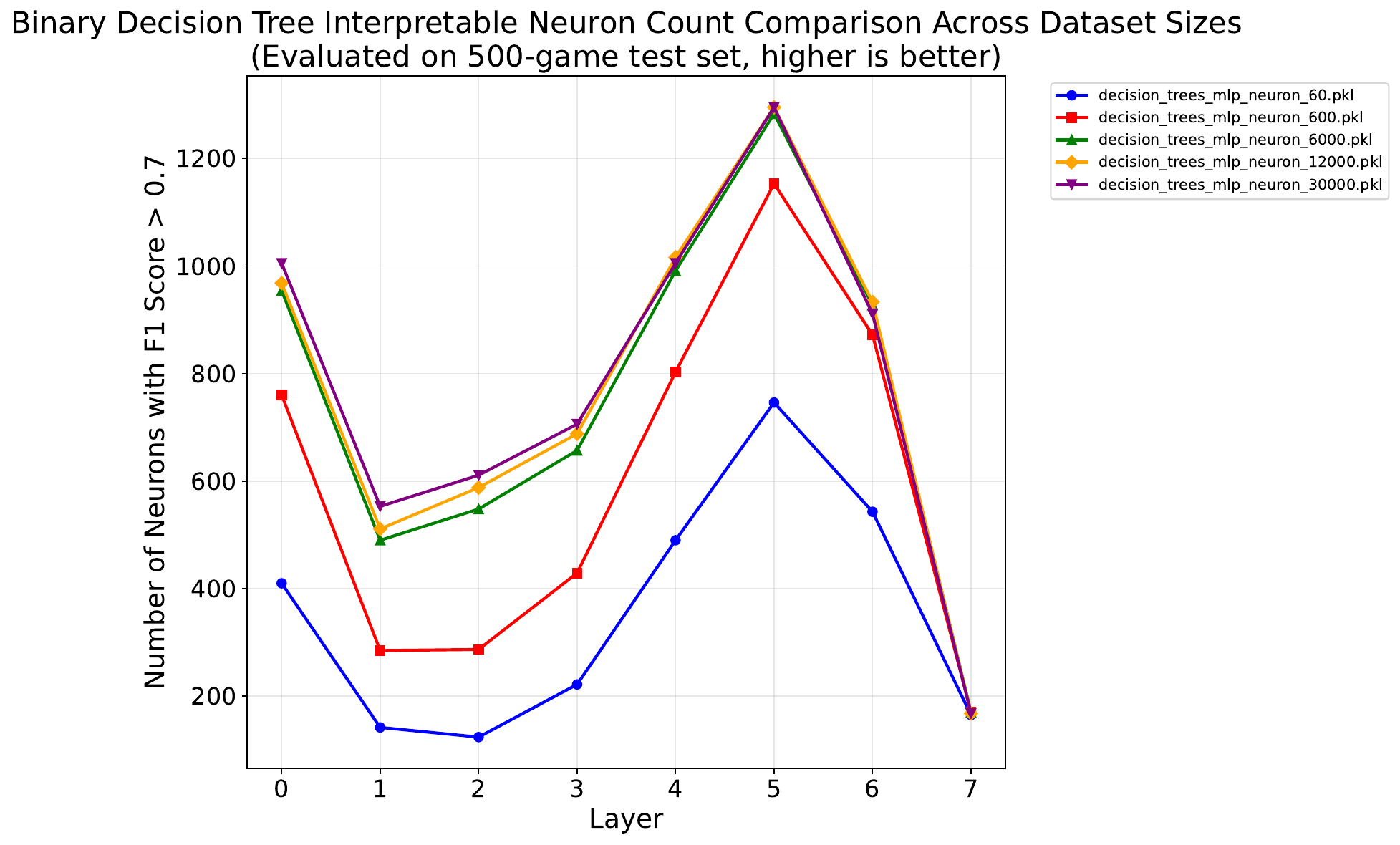}
  \end{center}
  \caption{Number of interpretable neurons with F1 score > 0.7 for depth 8 binary decision trees trained with different size of datasets.}\label{Fig:probe-features}
\end{figure}

\newpage
\section{Probe feature extraction}\label{probe_feature_extraction}
We computed the cosine similarity between each neuron’s input weights and the five probe directions: Mine, Empty, Theirs, Flipped, and Just Played. We then identified probe activation features with notably large or small values by filtering those exceeding two standard deviations from the mean.

\begin{figure}[!htbp]
  \begin{center}
    \includegraphics[width=1\textwidth]{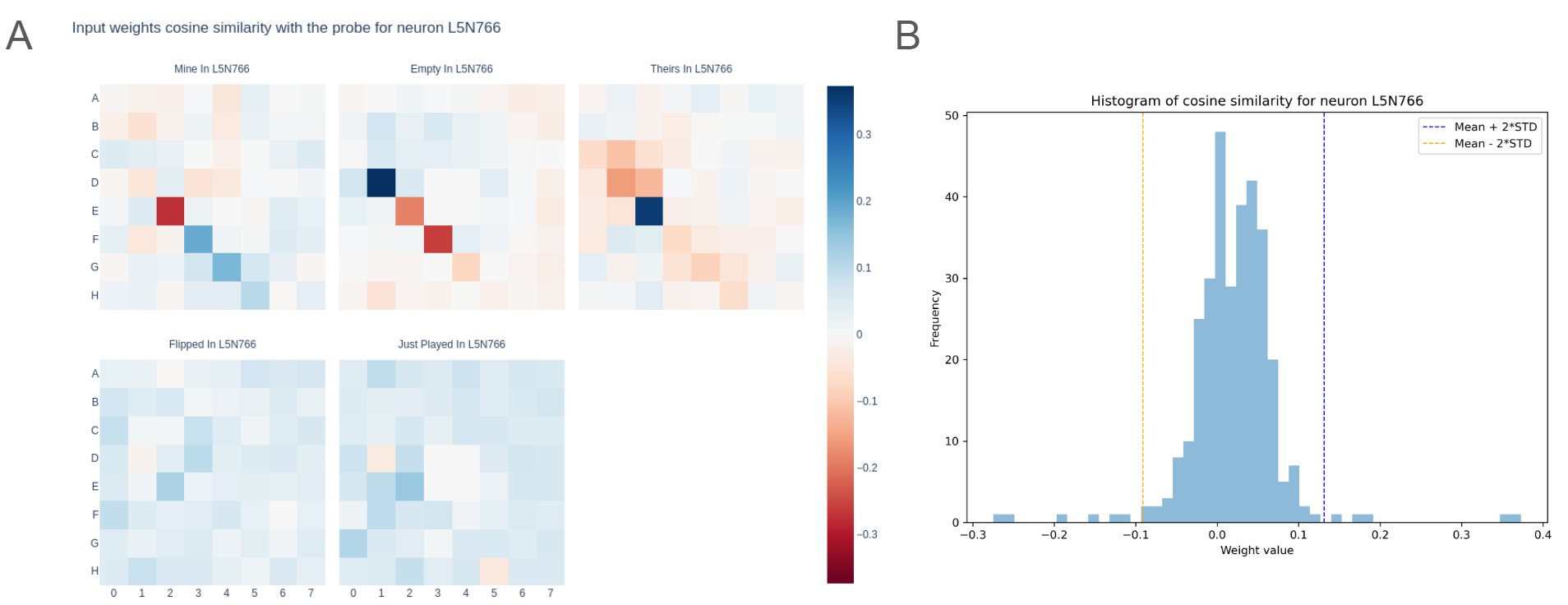}
  \end{center}
  \caption{Probe features. A. Cosine similarity between neuron input weights and Mine, Empty, Theirs, Flipped, Just Played probe directions. B. Distribution of the similarity values}\label{Fig:probe-features}
\end{figure}

\newpage
\section{RIPPER Feature Scoring and Selection}
\label{app:ripper_features}

For each neuron, RIPPER produces an ordered list of conjunctive rules.  
We assign each feature \(f\) a weight
\[
w(f) = \sum_{r:\,f\in r} 
      F^{(n)}_{1,\mathrm{strong}}
      \times \frac{1}{|r|}
      \times \frac{1}{\mathrm{rank}(r)^{\rho}},
\]
where \(F^{(n)}_{1,\mathrm{strong}}\) is the neuron’s F1-score on the positive class, \(|r|\) is the rule length, and \(\rho=0.7\) controls decay with rule rank.  

Top RIPPER features are then selected by applying a \(k\)-sigma filter (\(k=2\)) on these weights, retaining only those whose scores exceed the mean by at least two standard deviations.  
The resulting feature sets are used in the containment analysis in Sec.~\ref{metric-results}.



\newpage
\section{Changes of logits of valid move neuron zero ablation experiment}\label{logits_change_ablation}
We are showing three different possible logit differences before and after ablation. In the first case, the logit of D1 dropped to 4.0452 from 8.5554 but still significantly higher than the other illegal tokens (2.1943). In the second case, the logit of D1 dropped to 2.5962 that is close to the other illegal tokens (2.2567). In these two cases, the logit of D1 is still within the top ``valid move number'' of each play. So, we consider these two as accurate prediction. In the third case, the logit of D1 dropped to 1.6110 which is below some other illegal tokens and also out of 8 (``valid move number'' of this play). So, we consider these scenario as inaccurate prediction.

\begin{figure}[!htbp]
  \begin{center}
    \includegraphics[width=1\textwidth]{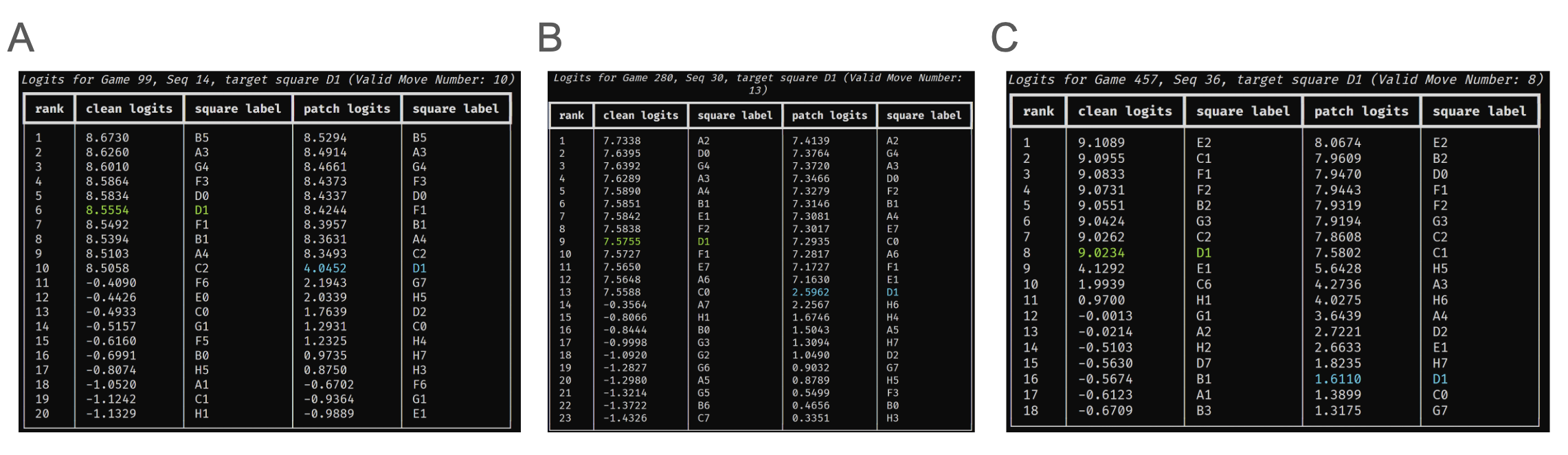}
  \end{center}
  \caption{Output logits of the model with or without ablation.}\label{Fig:logits-difference}
\end{figure}

\newpage
\section{Uniqueness of board state representation}\label{retrain_probe}
Prior work has shown that OthelloGPT linearly represents board state, and moreover, that this representation is causal \cite{nanda_emergent_2023}. However, this does not answer the question of whether there exist redundant representations of board state, or if the representation is unique. To test this, we first train a probe $P \in \mathbb{R}^{d_{model} \times 3}$ for each square on the residual stream at the end of layer $5$. We train over a train set of 50k games, and the probe achieves an accuracy of 99.4\% on a test set of 10k games. We then ablate the probe directions and retrain the probe. Specifically, a three class classification probe is parameterized by two directions $\{\Delta_1, \Delta_2\}$ corresponding to any two of the difference directions of the probe. We zero out the components of the activations in span($\{\Delta_1, \Delta_2\}$), and retrain the probe over the same train set. Evaluated over the same test set, the ablated probe performs as good as random with 33\% accuracy, confirming uniqueness of the board state representation.

\begin{figure}[!htbp]
  \begin{center}
    \includegraphics[width=1\textwidth]{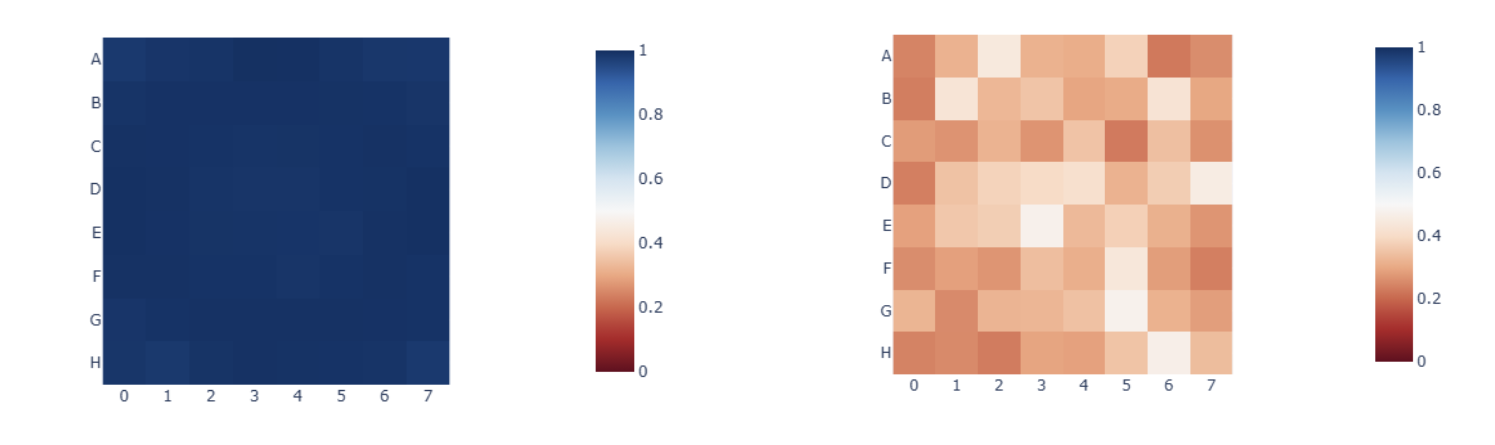}
  \end{center}
  \caption{Left: Original Probe Accuracy. Right: Ablated Probe Accuracy.}\label{Fig:ablated-probe}
\end{figure}

\newpage
\section{G2 was flipped decision tree neurons}\label{G2_flipped_dt}
\begin{figure}[!htbp]
  \begin{center}
    \includegraphics[width=1\textwidth]{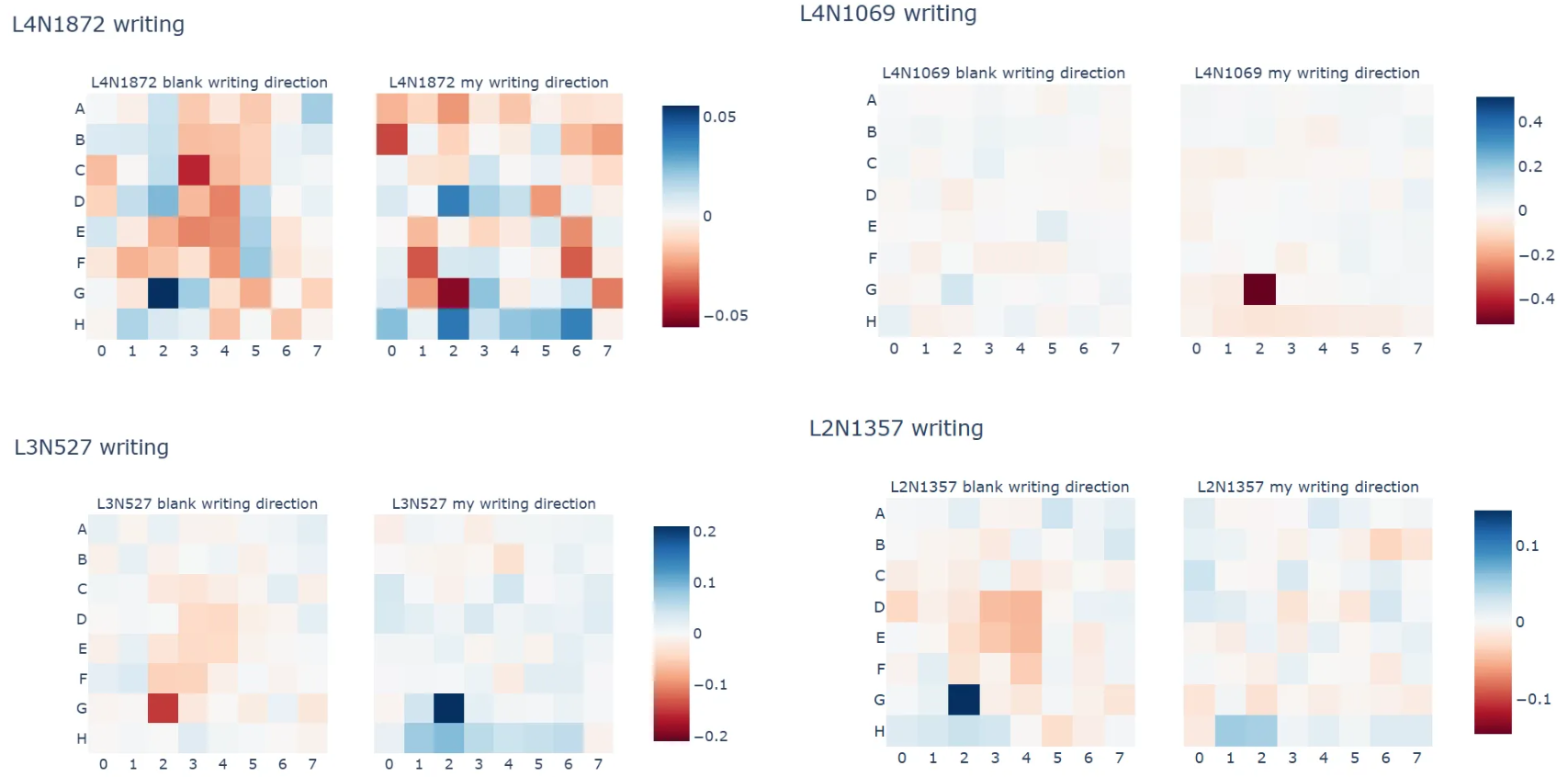}
  \end{center}
  \caption{Top neurons by F1 for G2 is flipped}\label{fig:G2_flipped_dt_neurons}
\end{figure}

While direct probe attribution filters for neurons that actively write out G2 is theirs when G2 is flipped, sorting decision trees by F1 only biases towards neurons that exclusively fire when G2 is flipped, not necessarily that they write out G2 is theirs. By visualizing heatmaps of cosine similarities of decoder weights and probe directions for the top four neurons by F1 (Figure~\ref{fig:G2_flipped_dt_neurons}), we see that while the first two write out G2 is theirs, the second two carry out different functions related to G2. 

\begin{figure}[!htbp]
  \begin{center}
    \includegraphics[width=1\textwidth]{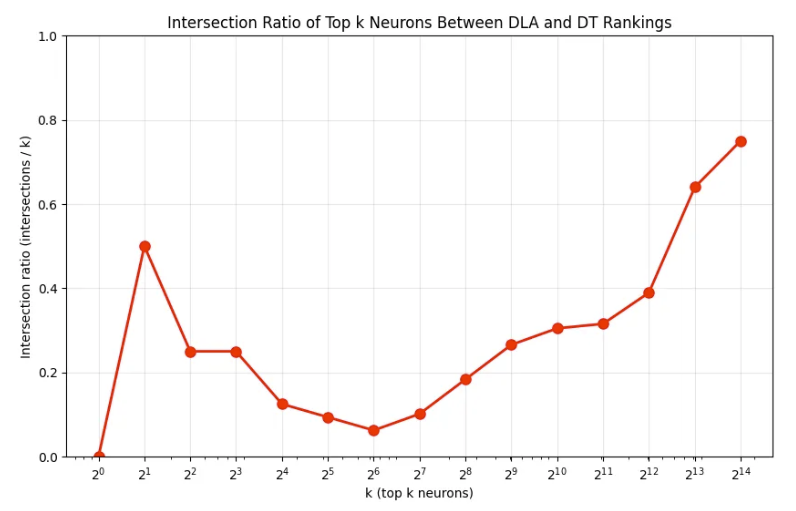}
  \end{center}
  \caption{Intersection ratio of top-$k$ neurons between direct probe attribution and decision tree F1 rankings for G2 is flipped.}\label{fig:G2_flipped_dpa_neurons}
\end{figure}

\newpage
\section{Fine-grained intervention set-up details}\label{sec:intervention_confounding}
To ensure the intervention and control conditions are not confounded, we select the intervention and control 3 square patterns such that the Manhattan distance of the two patterns to the middle 2x2 are equal. This corresponds to one diagonal pattern pointing to the middle, and an axial direction (either row or column), depending on the location of the square on the board. We also randomize assignment of the diagonal and axial pattern to intervention and control.

\end{document}